\documentclass{article}

\usepackage[preprint]{neurips_2024}
\usepackage{mdframed}
\usepackage{graphicx}
\usepackage{float}
\usepackage{subfigure}
\usepackage{bbding}
 
\usepackage[utf8]{inputenc} % allow utf-8 input
\usepackage[T1]{fontenc}    % use 8-bit T1 fonts
\usepackage{hyperref}       % hyperlinks
\usepackage{url}            % simple URL typesetting
\usepackage{booktabs}       % professional-quality tables
\usepackage{amsfonts}       % blackboard math symbols
\usepackage{nicefrac}       % compact symbols for 1/2, etc.
\usepackage{microtype}      % microtypography
\usepackage{xcolor}         % colors

\usepackage{amsthm,amsmath,amssymb}
\usepackage{multirow}
\usepackage{enumerate}
\usepackage{enumitem}

\usepackage{algorithm}
\usepackage{algorithmic}
\usepackage{hyperref}
\usepackage{hyperref}

\usepackage{wrapfig}

\newtheorem{principle}{Principle}
\newtheorem{lemma}{Lemma}

\title{Less is More: Pseudo-Label Filtering for\\Continual Test-Time Adaptation}

\author{
    Jiayao Tan\textsuperscript{\rm 1}\thanks{Co-first authors.} ,
    Fan Lyu\textsuperscript{\rm 2}\footnotemark[1] ,
    Chenggong Ni\textsuperscript{\rm 1},
    Tingliang Feng\textsuperscript{\rm 3}, \\
    \textbf{Fuyuan Hu\textsuperscript{\rm 1}\thanks{Corresponding author.} ,
    Zhang Zhang\textsuperscript{\rm 2},
    Shaochuang Zhao\textsuperscript{\rm 4},
    Liang Wang\textsuperscript{\rm 2}} \\
    \textsuperscript{\rm 1}School of Electronics and Information Engineering, Suzhou University of Science and Technology
    \\ \textsuperscript{\rm 2}New Laboratory of Pattern Recognition, Institute of Automation, Chinese Academy of Sciences 
    \\ \textsuperscript{\rm 3}College of Intelligence and Computing, Tianjin University
    \\ \textsuperscript{\rm 4}School of Electronics and Information Engineering, Jiangnan University
    \\  \texttt{\{jiayaotan@post, chenggongni@post, fuyuanhu@mail\}.usts.edu.cn, }\\\texttt{fan.lyu@cripac.ia.ac.cn, \{zzhang, wangliang\}@nlpr.ia.ac.cn,}
    \\ \texttt{fengtl@tju.edu.cn, zsc960813@163.com}
}

\begin{document}

\maketitle

\begin{abstract}
Continual Test-Time Adaptation (CTTA) aims to adapt a pre-trained model to a sequence of target domains during the test phase without accessing the source data. To adapt to unlabeled data from unknown domains, existing methods rely on constructing pseudo-labels for all samples and updating the model through self-training. However, these pseudo-labels often involve noise, leading to insufficient adaptation.
To improve the quality of pseudo-labels, we propose a pseudo-label selection method for CTTA, called Pseudo Labeling Filter (PLF).
The key idea of PLF is to keep selecting appropriate thresholds for pseudo-labels and identify reliable ones for self-training.
Specifically, we present three principles for setting thresholds during continuous domain learning, including initialization, growth and diversity. Based on these principles, we design Self-Adaptive Thresholding to filter pseudo-labels. Additionally, we introduce a Class Prior Alignment (CPA) method to encourage the model to make diverse predictions for unknown domain samples. Through extensive experiments, PLF outperforms current state-of-the-art methods, proving its effectiveness in CTTA.
Our code is available at \url{https://github.com/tjy1423317192/PLF}.
\end{abstract}

\section{Introduction}
\label{Introduction}
In recent years, Test-Time Adaptation (TTA) \cite{Domain1, Domain2, Domain3} has been well studied, which is a technique in unsupervised settings, aimed at adapting models trained on a source domain to new target domains without accessing the source data during inference. However, TTA is insufficient in many real-world applications like autonomous driving, where sensors are influenced by various factors such as weather, lighting conditions, and traffic situations \cite{traffic}.
% , leading to differs from traditional TTA, CTTA continuously adjusts the model during inference to account for the evolving data distribution of the target domain. 
CTTA~\cite{Cotta} is proposed as an online continuous form of TTA.
CTTA aims to continually adjust models to adapt to the evolving data distribution of target domains without relying on source data. This enables models to continuously adapt \cite{continullearning, continullearning2, continullearning3, continullearning4} to new environments in many practical unsupervised settings.

Currently, the mainstream CTTA methods highly rely on \textit{pseudolabeling} techniques, which perform well in short-term domain adaptation under TTA and are widely adopted. Specifically, methods based on Mean-Teacher (MT) structure yielded impressive results. 
For samples from the target domain, MT generates pseudo-labels \cite{pseudolabel1,pseudolabel2} and optimizes the loss of consistency between the pseudo-labels and the predictions, thus facilitating model adaptation to the target domain \cite{aug1,aug2}. 
However, the pseudo-label-based approach faces significant challenges in long-term CTTA applications. 
Over time, the presence of \textit{low-quality} pseudo-labels may lead to the accumulation of errors in the testing process, which can lead to performance degradation. 
% And existing methods lack effective mechanisms to address this challenge. 
% Inspired by this, we believe that pseudo-label filtering is a crucial step. 
Thus, it is reasonable to filter out these low-quality pseudo-labels in CTTA.
Setting a threshold is a natural and straightforward method for pseudo-label filtering.
% can be set to filter high-quality pseudo-labels and reduce the negative effects generated by pseudo-labels. 
However, a fixed threshold by fine adjusting is impractical in unsupervised CTTA tasks, where the test data is unavailable and the domain shift is unknown in advance.

% due to the continuity and on-linearity of CTTA data features, it is impossible to directly access all the data, and the simple method of setting a fixed threshold for each domain by directly setting a fixed threshold for each domain is not reliable in testing scenarios.
% Currently, the mainstream methods of CTTA still rely on pseudo-labeling techniques, which have shown excellent performance in short-term domain adaptation under TTA and are widely adopted. Among them, Mean Teacher (MT) \cite{Cotta} stands out as a representative approach. MT generates pseudo-labels \cite{pseudolabel1,pseudolabel2} for target domain samples and optimizes consistency loss between pseudo-labels and ground truth labels, thereby facilitating model adaptation to the target domain \cite{aug1,aug2}. However, in long-term CTTA applications, pseudo-labeling-based methods face significant challenges. Firstly, as time progresses, the continuous changes in the data distribution of the target domain lead to a gradual decline in the accuracy of pseudo-labels, thereby impacting model performance. Secondly, the presence of low-quality pseudo-labels \cite{incorr} may result in error accumulation during the training process, leading to performance degradation and issues such as model forgetting.
\begin{figure}
  \centering
  \includegraphics[width=0.9\linewidth]{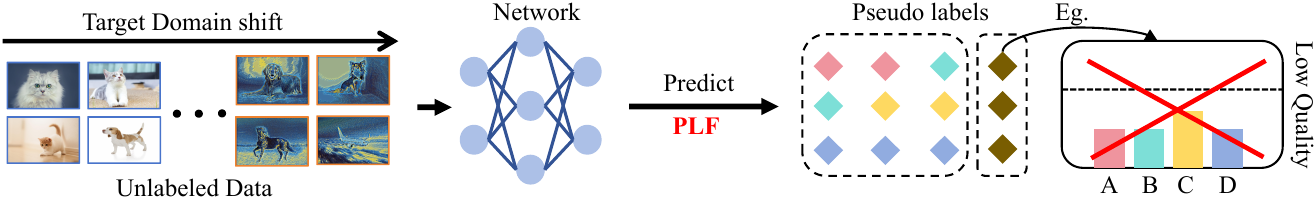}
  \caption{The pre-trained network generates pseudo-labels for given samples. Some unlabeled samples may be given the wrong pseudo-labels (dark color). Therefore, treating all samples equally will lead to error accumulation. PLF introduces three principles to set an adaptive threshold to monitor the generated pseudo-labels so that the correct pseudo-label (light color) is selected for learning.}
  \label{cou learn}
  % \vspace{-10px}
\end{figure}
% The core reason for these challenges in long-term CTTA lies in the gradual decline in the reliability of pseudo-labels, with existing methods lacking effective mechanisms to address this challenge. Inspired by this, we believe that pseudo-label filtering is a crucial step. By setting thresholds to filter out high-quality pseudo-labels, the adverse impact of low-quality pseudo-labels on model performance can be reduced. However, due to the continuity and variability of CTTA data features, pseudo-label filtering is not straightforward. Although the simplest approach is to set a fixed threshold for each domain, this method is not feasible in testing scenarios where all data cannot be directly obtained due to the linear data flow.

% Therefore, exploring how to identify optimal thresholds to filter pseudo-labels for addressing unknown domain shifts is a worthwhile endeavor. 
In this paper, we explore how to keep adjusting thresholds for pseudo-label filtering in testing scenario with continuous domain shifts, and we propose three principles for setting pseudo-label thresholds in the CTTA task:
% To this end, we derive three guiding principles for designing thresholds through illustrative examples. 
(1) \textit{Low-level initialization}: At the early testing stages, thresholds are suggested to be relative small to encourage diverse pseudo labels, improve unlabeled data utilization, and fasten convergence.
(2) \textit{Positively correlated to confidence}: The CTTA process is intricate, and it's optimal to dynamically adjust thresholds while maintaining a positive correlation with model confidence. This ensures stable, high-quality pseudo-label learning throughout the process.
(3) \textit{Vary in different classes}: 
Throughout the CTTA process, the states of various categories in the adaptation target domains exhibit variability. Hence, it's beneficial to implement fine-grained, class-specific thresholds to ensure equitable assignment of pseudo labels across different classes.

Based on the principles, we propose a pseudo-label selection method, termed Pseudo Labeling Filter (PLF), which builds self-adaptive thresholds capable of accommodating CTTA process to ensure the quality of pseudo-labels.
Specifically, we establish initial thresholds and class-level thresholds tailored to the number of classes.
Then, we utilize Exponential Moving Average (EMA) \cite{EMA} and Exponential Decay (ED) strategy on the confidence scores of unlabeled data to ensure a positive correlation between thresholds and the model confidence for updating adaptive thresholds. 
Moreover, considering that the difficulty of class learning varies across domain transformations, we propose a class-balanced regularization objective, encouraging the model to generate diverse predictions across all classes, thereby reducing error accumulation caused by continuous domain shifts \cite{shifts1,shifts2,shifts3,continullearning5}. Our contributions are as follows:
% 为了进一步优化类阈值使模型在CTTA过程中的多样化预测
\begin{enumerate}[label=(\arabic*),left=0pt,itemsep=0pt]
    \vspace{-5px}
    \item We derive three principles for setting pseudo-label thresholds in the CTTA task, including low-level initialization, positively correlated to confidence and vary in different classes.
    \item We propose PLF based on the three principles, selecting high-quality pseudo-labels for adaptation, and effectively reducing the negative impact of error accumulation of CTTA.
    \item We introduce Class Prior Alignment to encourage diverse predictions on samples from continuous domains, reducing the risk of propagating misleading information during the CTTA task.
\end{enumerate}

\section{Related Work}
\label{Related Work}
\subsection{Continual Test-time Domain Adaptation}
To address the challenge of CTTA, various solutions have been proposed building upon TTA~\cite{shot,TTT,TTT2,TTT3}. The online version of TENT~\cite{TENT}, is a viable approach, updating the trainable batch normalization parameters of pre-trained models at test time by minimizing the entropy of model predictions. AdaContrast \cite{AdaContrast} employs contrastive learning with online pseudo refinement to learn better feature representations, reducing noisy pseudo-labels. Conjugate PL \cite{Conjugate} presents a general approach for obtaining test-time adaptation loss, enhancing robustness to distribution shifts. However, a drawback of early methods is they never consider the continuous domain shifts.
A significant advancement in this field is the development of Continual Test-Time Adaptation (Cotta) \cite{Cotta}. Cotta is the first method explicitly tailored to the demands of CTTA. Cotta adopts a weighted augmentation-averaged mean teacher framework, drawing insights from prior work such as mean teacher predictions introduced by Tarvainen and Valpola \cite{Meanteacher}. 
% The student-teacher framework proposed by Cotta \cite{Cotta} serves as a foundational architecture for numerous subsequent studies. 
Niu et al. \cite{Niu} adopt a similar student-teacher framework, integrating continuous batch normalization statistics updates to reduce computational costs and enhance adaptation efficiency. Another notable approach explored by researchers is leveraging mean teacher settings for symmetric cross-entropy and contrastive learning, as demonstrated in RMT~\cite{RMT}.

\subsection{Pseudo-Label Filtering}
Pseudo-labeling is a widely adopted technique in self-learning, where the model's output class probabilities are utilized as training labels. FixMatch \cite{fixmatch} generates pseudo-labels based on the model predictions for weakly augmented unlabeled images, then the model is trained to match the pseudo-labels with predictions on strongly augmented images.
Softmatch \cite{softmatch} not only generates reliable pseudo-labels with high confidence and low uncertainty but also incorporates thresholding techniques to further reduce model calibration errors.
% In the context of unsupervised domain adaptation, pseudo-labeling typically involves generating labels for unlabeled target samples using the predicted class probabilities of the source model. 
Existing methods primarily rely on pseudo-labels as a form of ``supervision'' to compensate for the lack of ground truth labels in the target domain. However, they have not closely examined the quality of pseudo-labels \cite{quality1,quality2}, as the use of mislabeled samples in self-learning accelerates error accumulation. 
Recently, DSS \cite{DSS} proposes joint positive and negative learning with dynamic threshold modules to minimize error accumulation \cite{error1,error2,error3} from mislabeled pseudo-labels. However, DSS does not fully consider the matching trends of thresholds in the continuous domain adaptation process and the importance of maintaining class balance in pseudo-label selection. 
In this paper, we introduce an adaptive cross-domain threshold mechanism based on joint class learning states to minimize the error accumulation effect from mislabeled pseudo-labels and design a Class Prior Alignment mechanism to encourage the model to make diverse predictions \cite{diverse1,diverse2,diverse3} for samples in continuous domains, thereby reducing error accumulation caused by cross-domain shifts.

\section{Principles of Thresholding in CTTA}
\label{Principles}
\subsection{Problem definition}
In CTTA, we utilize an MT framework for an existing pre-trained model $f_\theta$ parameterized on the source data $(\mathcal{X}^\text{S}, \mathcal{Y}^\text{S})$, where $\mathcal{X}^\text{S}$ and $\mathcal{Y}^\text{S}$ are the data sets and the corresponding label sets.
After trained well, CTTA aims to make the model in response to the continually changing target domain $\mathcal{X}^D$ during test phase, without accessing any source data, here we simplify $\mathcal{X}^D$ to $\mathcal{X}$.
% Unlabeled target domain data is sequentially provided, with the model only having access to the data at the current time step. 
We achieve this by dynamically filtering pseudo-labels in the MT framework, where the pseudo-labels are generated by a teacher model $f_\text{t}(\cdot)$ in an online manner. 
To facilitate the description of the equations later, we use $q$ and $Q$ to denote the abbreviation of student model's prediction $f_\text{s}(\mathcal{X})$ and teacher model's prediction $f_\text{t}(\mathcal{X})$. In this section, we draw inspiration from FreeMatch \cite{freematch} to outline the principles of establishing adaptive thresholds that meet CTTA criteria using multi-class classification examples.

% In CTTA, Given an existing pre-trained model $f_\theta$ with parameters $\theta$ trained on the source data $(\mathcal{X}^\text{S}, \mathcal{Y}^\text{S})$, we aim at improving the performance of this existing model during inference time for a continually changing target domain in an online fashion without having access to any source data. Unlabeled target domain data $X^T$ is provided sequentially and the model only has access to the data of the current time step.

\subsection{Three Principles of Thresholding in CTTA}
\label{sec:principle}
For any test domain $\mathcal{X}$, we assume that input logits for each class
% $f_\theta(X) \in \{l_1,l_2,...,l_c\}$ 
meets a Gaussian distribution:
\begin{equation}
l_c \sim N(\mu_c,\sigma_c^2),\quad \forall c \in \mathcal{C},
\end{equation}
% Assuming without loss of generality that $\mu_1 > \mu_2$, the confidence score s(x) output by the classifier can be expressed as $s(x) = 1/[1 + exp(- \beta (x-(\mu_1 + \mu_2 )/2))]$, where $\beta$ is a positive parameter reflecting the model's learning state. The model's confidence is influenced by continuous domain changes, exhibiting a stepped growth rather than a smooth one. In fact, $(\mu_1 + \mu_2)/2$ serves as the optimal linear decision boundary in Bayesian terms. We consider a scenario where a fixed threshold $\tau \in (1/2,1)$ is used to generate pseudo-labels. If $s(x) > \tau$, the sample x is assigned the pseudo-label +1; if $s(x) < 1 - \tau$, the pseudo-label is -1. If $1 - \tau \leq s(x) \leq  \tau$, the pseudo-label is 0 (masked).
where $\mathcal{C}$ is the class sets.
The confidence score can be then computed by ${\rm Softmax}(l_c) = e^{\beta{l_c}}/\sum\nolimits_{i=1}^{C}e^{\beta{l_i}}$, where $\beta$ is a positive scaling parameter.
% The model confidence is influenced by continuous domain changes, exhibiting a stepped growth rather than a smooth one. 

We first consider the case of generating pseudo-labels using a fixed threshold $\tau$. 
If ${\rm Softmax}(l) > \tau$ then the pseudo-label is retained, otherwise the pseudo-label is discarded.
{Inspired by FreeMatch \cite{freematch}, we then derive the following Lemma to show the principles of self-adaptive threshold:}

% \textbf{Theorem 3.1} \emph{For a binary classification problem as mentioned above, the pseudo label $Y$ has the following probability distribution:}
\begin{lemma}[\textbf{Class Probability Distribution (CPD)}]
\label{lemma}
For a multi-class classification problem as mentioned above, the probability distribution of the class with maximum confidence in the soft pseudo label $Y_p$ is as follows:
% For a multi-class classification problem as mentioned above, the pseudo label $Y_p$ has the following probability distribution:
\begin{equation}
\begin{aligned}
& P(X|Y_p = c) = \Phi\left(\frac{1}{\sigma_c}\left(\mu_c-\frac{1}{\beta}\log(\frac{\tau}{1-\tau})-\frac{1}{\beta}\log\sum\nolimits_{i \neq c} e^{\beta l_i}\right)\right), \\
& P(X|Y_p = 0) = 1-\sum\nolimits_{i}P({X|Y_p} = i),
\end{aligned}
\end{equation}
where $\Phi$ is the cumulative distribution function of a standard normal distribution.
\end{lemma}
% \begin{equation}
% \begin{aligned}
% & P({Y} = 1) = \frac{1}{2}\Phi (\frac{{\frac{{{\mu _{(2,D)}} - {\mu _{(1,D)}}}}{2} - \frac{1}{{{\beta }}}\log (\frac{{{\tau}}}{{1 - {\tau}}})}}{{{\sigma _{(2,D)}}}}) + \frac{1}{2}\Phi (\frac{{\frac{{{\mu _{(1,D)}} - {\mu _{(2,D)}}}}{2} - \frac{1}{{{\beta }}}\log (\frac{{{\tau}}}{{1 - {\tau}}})}}{{{\sigma _{(1,D)}}}}),\\
% & P({Y} = -1) = \frac{1}{2}\Phi (\frac{{\frac{{{\mu _{(2,D)}} - {\mu _{(1,D)}}}}{2} - \frac{1}{{{\beta }}}\log (\frac{{{\tau}}}{{1 - {\tau}}})}}{{{\sigma _{(1,D)}}}}) + \frac{1}{2}\Phi (\frac{{\frac{{{\mu _{(1,D)}} - {\mu _{(2,D)}}}}{2} - \frac{1}{{{\beta }}}\log (\frac{{{\tau}}}{{1 - {\tau}}})}}{{{\sigma _{(2,D)}}}}) \\
% & P({Y} = 0) = 1-P({Y} = 1) -P({Y} = -1) 
% \end{aligned}
% \end{equation}

% & P({Y} = 1) = P({Y} = -1) = \frac{1}{2}\Phi (\frac{{\frac{{{\mu _2^D} - {\mu _1^D}}}{2} - \frac{1}{\beta}\log (\frac{{{\tau}}}{{1 - {\tau}}})}}{{{\sigma _2^D}}}) + \frac{1}{2}\Phi (\frac{{\frac{{{\mu _1^D} - {\mu _2^D}}}{2} - \frac{1}{\beta}\log (\frac{{{\tau}}}{{1 - {\tau}}})}}{{{\sigma _1^D}}}),\\
% & P({Y} = 0) = 1-P({Y} = 1||Y = -1)
The Proof is shown in Appendix~\ref{proof}. 
Based on Lemma~\ref{lemma}, we can obtain the following principles:

% \begin{mdframed}
\begin{principle}[Low-level initialization (Derivation in Appendix~\ref{proof-1})]
{The utilization rate of high-quality data denoted as $1 - P({Y} = 0)$ is directly controlled by the threshold $\tau$.
% see proof.\ref{proof-1} for specific certificates.
As $\tau$ increases, the data utilization rate decreases. During the test phase, since $\beta$ remains small, adopting a high threshold may result in a low sampling rate and slow convergence.}
\label{principle1}
\end{principle}
% \end{mdframed}

% \begin{mdframed}
\begin{principle}[Positively correlated to confidence (Derivation in Appendix~\ref{Derivation2})]
{Fluctuations in the sample distribution can cause $\beta$ to show nonlinear growth due to domain shifts, which affects the utilization of high-quality data $1-P({Y}=0)$. 
% shown in proof.\ref{proof-2}
To ensure the stability of the sampling rate during CTTA, a positive correlation between $\beta$ and $\tau$ can be maintained by maintaining $\beta$.}
\label{principle2}
\end{principle}
% \end{mdframed}
% Therefore, the threshold needs to be adaptively adjusted to accommodate these changes. Otherwise, high-quality samples may be erroneously filtered out, leading to insufficient adaptation of the model to the new domain.

% \begin{mdframed}
\begin{principle}[Vary in different classes (Derivation in Appendix~\ref{Derivation3})]
{The utilization of high-quality data $1 - P({Y} = 0)$ decreases as the distance between $\mu$ decreases. During the CTTA process, certain categories may show closer category centroids due to domain bias, making it more likely that unlabeled samples will be incorrectly predicted. Therefore, fine thresholds are set for pseudo-labeling filtering to ensure high sampling rates.}
\label{principle3}
\end{principle}
% \end{mdframed}
% Hence, a moderate threshold is needed to balance the sampling rate. Otherwise, we may not have enough samples to train the model to distinguish between classes in the new domain. In other words, as the two classes become more similar due to domain changes, an unlabeled sample is more likely to be masked. With the increasing similarity between these classes, more mixed samples are in the feature space, and the model has less confidence in its predictions. Hence, a moderate threshold is needed to balance the sampling rate. Otherwise, we may not have enough samples to train the model to distinguish between classes in the new domain.

\subsection{Discussion}
First, to verify the validity of Principle \ref{principle1}, we set different initial thresholds to observe the model's adaptation to the target domain. The results (Table.\ref{ini}) show that a lower initial threshold helps to improve the model effect. Therefore, we believe it is beneficial to set lower initial thresholds in the early training phase. This practice encourages the generation of more pseudo-labels, which improves the utilization of unlabeled data and fasten convergence.

Secondly, to assess the validity of Principle \ref{principle2}, we design different ways of matching thresholds to model confidence for target domain testing, including fixed thresholds and multiple positive correlation matching methods. The results (Table.\ref{Matching and Granularity}) show that fixed thresholds lead to unacceptable confirmation bias, while the matching approaches that keep the thresholds positively correlated with model confidence perform well. Therefore, we believe that ideally, thresholds should be positively correlated with model confidence regardless of domain variations to maintain a stable sampling rate.

Finally, to verify the theoretical value of Principle \ref{principle3}, we conduct experiments on coarse-grained and fine-grained thresholding. The results
(Table. \ref{Matching and Granularity}) show that fine-grained thresholding performs better. Therefore, we believe that in CTTA, the use of coarse-grained thresholds makes it more difficult to distinguish between categories due to more similarities. Therefore, it is necessary to use category-specific fine-grained thresholds to fairly assign pseudo-labels to different categories. However, based on the three principles mentioned above, how to combine MT for effective implementation in CTTA is the remaining issue.

% The three principles of threshold adjustment during CTTA can be summarized as follows. Firstly, due to the poorer learning state of the early model, an initial lower threshold is needed to encourage the generation of diverse pseudo-labels, thereby enhancing the utilization of unlabeled data. Secondly, as training progresses, a fixed threshold would result in unacceptable confirmation bias. Ideally, regardless of domain variations, the threshold $\tau$ should change positively with $\beta$ to maintain a stable sampling rate. Lastly, as domain changes may reduce differences in class data distributions, making it harder to distinguish between classes, a fine-grained class-specific threshold is required to fairly allocate pseudo-labels to different classes. The challenge lies in designing a threshold adjustment scheme that comprehensively considers all these factors, thereby reducing sample selection errors caused by domain shifts and minimizing the risk of error accumulation.

\section{Methodology}
\label{Methodology}
\subsection{OverView}
\begin{figure}
  \centering
  \includegraphics[width=\linewidth]{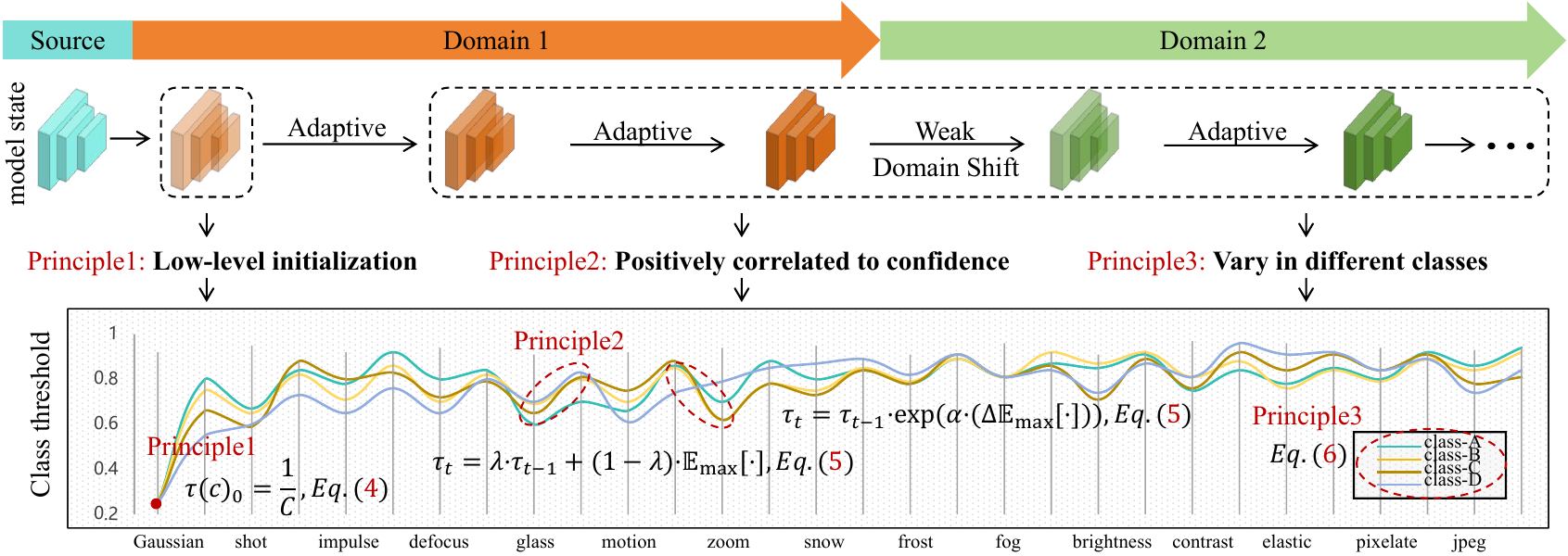}
  % \caption{The proposed Pseudo-Label Filtering (PLF) framework for continual test-time domain adaptation (CTTA). Unlabeled target domains evolve. At time step t, the student model initially predicts the sample $X_t$ from domain $D$. Subsequently, the teacher model retrieves input-enhanced images $X_t$ from domain $D$ and filters the pseudo-labels $Y_t$ predicted by the teacher based on the threshold settings under Principles, for cross-entropy loss computation. Finally, Class Prior Alignment (CPA) is calculated using the uniform prior distribution coefficients generated from the predictions of the student and teacher models to balance inter-class predictions.}
  \caption{The proposed Self-Adaptive Thresholding for CTTA. 
The different colors of the model represent adapted model parameters for various domains, with the depth of color indicating the degree of adaptation to each domain. Based on the threshold design principles corresponding to each stage of CTTA, high-quality pseudo-labels can be obtained. Additionally, we plot an adaptive curve in CIFAR10-C for the thresholds to better observe their changing trends during the CTTA process.}
  \label{model}
\end{figure}

% We believe that the key to setting thresholds in the CTTA process is that the thresholds should reflect the model confidence in the target domain, as shown in Fig.\ref{model}.
% The learning effect can be estimated by the prediction confidence of a well-calibrated model.
Based on the above principles, we propose two parts of the Pseudo-Label Filtering method: the \textbf{Self-Adaptive Thresholding} (SAT) method and the \textbf{Class Prior Alignment} (CPA). SAT automatically defines and adaptively adjusts the confidence thresholds using the model prediction during the test phase, regardless of whether the target data domain has changed or not. In addition, to more effectively deal with class conflicts due to unlabeled settings, we propose CPA that encourages the model to generate diverse predictions across all classes, thereby reducing error accumulation due to domain shift. 
Our methods are illustrated in Fig.~\ref{model} and Algorithm \ref{alg}. 

% Furthermore, a fairness objective $L_f$ is typically introduced to encourage the model to predict each class at the same frequency, usually taking the form of $L_f = U(C) \log \mathbb{E}_B[q_s]$, where $U$ is a uniform prior distribution. However, using a uniform prior not only enables adaptation to non-uniform data distributions caused by domain shifts but also acknowledges the potential imbalance in pseudo-label distributions within the same target domain. Therefore, achieving fairness in utilizing samples with per-class thresholds across batches is crucial.

\subsection{Self-Adaptive Thresholding}
To sum up, according to Section \ref{sec:principle}, 
% establishing an adaptive threshold for the adaptive CTTA process ensures that potentially correct samples are adequately trained, thus facilitating fast convergence.:
we propose to set pseudo-label thresholds as follows.
\begin{equation}\label{3}
    \tau_t^*(c) = {\rm MaxNorm}(\tau_t(c)) \cdot  \tau_t = \frac{\tau_t(c)}{{\rm max}({\tau_t(c))}}\cdot\tau_t, \quad t\neq 0, \forall c \in \mathcal{C}
\end{equation}
where MaxNorm is the Maximum Normalization (i.e., $x' = \frac{x}{\max(x)}$), $\tau_t^*(c)$ represents the class-level adaptive threshold that combines the global threshold $\tau_t$ and class thresholds $\tau_t(c)$.
Where the global threshold reflect overall model confidence and the class thresholds reflect class-specific confidence.
% By Principle \ref{principle1}, we set a low threshold at the beginning of Testing to accept more potentially correct samples. Based on Principle \ref{principle2}, we design the threshold evolution function to ensure that a positive correlation is maintained between the model's learning state and the threshold as the target domain changes. Principle \ref{principle3} involves the formulation of class thresholds to modulate the global thresholds, which are then integrated to arrive at the final adaptive thresholds. 
In the following, we describe the design process of how to design initial threshold, global threshold, and class thresholds using three principles.

% We analogize the overall confidence of the model towards unlabeled data as the model's self-confidence and initialize the base threshold as the average confidence of unlabeled data. Firstly, We initialize the global threshold $\tau_t$ to $\frac{1}{C}$ according to Principle \ref{principle1}, where $C$ represents the number of categories (See the appendix for inspiration for setting $\frac{1}{C}$) :

\textbf{Low-level initialization.} 
% First, by analyzing the effects of different initial thresholds on the initial model adaptation to the target domain, we find that high thresholds (eg. $\tau@0.8$) have a certain effect on the pseudo-label filtering of the CTTA process, so based on Principle \ref{principle1}, we exclude the use of high initial thresholds and thus choose low initial thresholds related to the number of categories $\frac{1}{C}$. where $C$ stands for the number of categories 
First, by analyzing the effects of different initial thresholds on the initial model adaptation to the target domain (See Appendix \ref{Initial} for initial threshold comparison experiments), we find that high thresholds (e.g., $\tau@0.8$) have an effect on the pseudo-label filtering of the CTTA process. Therefore, according to Principle \ref{principle1}, we believe that a relatively small initialization threshold is useful.
There exist multiple selection strategies, we try to use several low thresholds for our experiments, and we find that all of them achieve better performance but are still diverse.
Since it is not practical to adjust the initialization value in the testing phase. We believe that choosing a low initialization related to the number of categories is a simple and effective way to set up the threshold:
\begin{equation}  
\label{4}
\tau_0^*(c) = \frac{1}{C}.
\end{equation} 
This can also be finely tuned in specific tasks (See Appendix \ref{initialization} for a theoretical derivation).
% Then, due to the nature of the online data stream of CTTA, it is impractical to compute the confidence for all unlabeled data at each time step. Therefore, based on Principle \ref{principle2}, we estimate the base confidence as the exponential moving average (EMA) of the confidence at each training time step. In addition, since the model may undergo domain changes at a given time step, the threshold needs to be penalized to match the confidence of the model. Therefore, we penalize the thresholds using the Exponential Decay (ED) algorithm. The thresholds are defined and adjusted as follows:

\textbf{Positively correlated to confidence.} 
During the CTTA process, the fluctuation of model confidence is unstable, with most instances showing an increase within the domain but a decrease across domains. Failure to adaptively adjust thresholds during this process may result in the erroneous filtration of numerous high-quality pseudo-labels. 
% To validate this, we employ fixed thresholds (eg. Fixed$\tau$@0.6, Fixed$\tau$@0.8), confirming its detrimental effects.
According to Principle \ref{principle2}, to match the thresholds with the confidence of the model, we use Exponential Moving Average \cite{EMA} (EMA) of the confidence based on each training time step as the intra-domain base confidence estimate and adjust the inter-domain thresholds using the Exponential Decay (ED) \cite{ED}. 
We also compare other algorithms that fit the positive correlation see Appendix \ref{pos} for comparative analysis.
Global thresholds are defined and adjusted as follows:
% In addition, due to the nature of CTTA's online data stream, it is impractical to compute confidence levels for all unlabeled data at every time step.
\begin{equation}  
\label{5}
\tau_t = 
\left\{  
     \begin{array}{ll}  
     \vspace{1ex}
    {\rm EMA}(Q_{t}) = 
     \lambda \cdot \tau_{t-1}+(1-\lambda)\cdot \mathbb{E}_{\max}[ {(Q_t)} ]
     , &\mathbb{E}_{\max}[Q_t]>\mathbb{E}_{\max}[Q_{t-1}],  \\
    {\rm ED}(Q_{t},Q_{t-1}) = \tau_{t-1}\cdot{\rm exp}(\alpha  (\mathbb{E}_{\max}[{Q_{t-1}}] - \mathbb{E}_{\max}[{Q_t}])), &{\rm otherwise}, \\  
     \end{array}  
\right.  
\end{equation}
we replace $\mathbb{E}_{\max}[\cdot]=\frac{1}{B}\sum\nolimits_{b=1}^{B}[{\rm max}(\cdot)]$, where $B$ Represents Batch size. 
$\lambda \in (0,1)$ is the momentum decay for EMA and $\alpha$ is the ED factor. 
EMA maintains a certain degree of smoothing while making the thresholds more sensitive to the latest changes, thus more effectively reflecting the actual trends in model confidence. 
ED introduces a nonlinear, decaying adjustment mechanism, making the threshold's response to performance degradation more robust and adaptive. 
When there is a significant difference in confidence, threshold adjustment will be substantial, and this design makes the threshold adjustment smoother, and less sensitive to minor performance degradation.

\textbf{Vary in different classes.} 
% Based on Principle \ref{principle3}, class thresholding aims to abstract global thresholds to a finer level of granularity, taking into account class diversity and potential class adjacencies to ensure sample availability. We compute the model's predicted expectation for each class $c$ to estimate the learning status of a particular class:
% By analyzing the \lyu{global} threshold CTTA experiments set based on the above principles we find that 
According to Principle \ref{principle3}, the confidence levels of different categories in the dynamic target domain show different trends and variations. 
Therefore, we try to use multiple thresholds for pseudo-tag filtering and find that although the number of fluctuations in the quality of pseudo-tags tends to stable, their overall number increases. 
This suggests that refining the thresholds to a finer granularity can accommodate the diversity between classes and potential class adjacencies, thus ensuring the usability of the samples. 
We compute the model's predicted expectation for each class $c$ to estimate the confidence of a given class:
\begin{equation}  
\label{6}
\tau_t(c) = 
\left\{  
     \begin{array}{ll}  
     \vspace{1ex}
     {\rm EMA}(Q_{t}(c))  &\mathbb{E}_{\max}[Q_t(c)]>\mathbb{E}_{\max}[Q_{t-1}(c)],  \\
    {\rm ED}(Q_{t}(c),Q_{t-1}(c))  &{\rm otherwise}, \\  
     \end{array}  
\right.  
\end{equation}
where $\tau_t(c)$ is the list containing all classes. 

Given the filtered pseudo-labels $\tau(c)$ for each class in Eq.~\eqref{3}, the objective $\mathcal{L}_{\rm u}$ is computed as:
\begin{equation} 
\label{7}
    \mathcal{L}_{\rm u} = \frac{1}{B}\sum\nolimits_{b=1}^{B}{\mathbf{1}({\rm max}(Q)>\tau^*(c))}\cdot \mathcal{H}(q,Q),
\end{equation}
where $\mathbf{1}(\cdot > \tau)$ is the indicator function for confidence-based thresholding with $\tau$ being the threshold. We focus on pseudo-labeling using symmetric cross-entropy loss $\mathcal{H}(\cdot)$ with confidence threshold for entropy minimization. 

In summary, by Principle \ref{principle1}, we set a low threshold at the beginning of testing to accept more potentially correct samples, as shown in eq\ref{4}. Based on Principle \ref{principle2}, we design the threshold evolution function to ensure that a positive correlation is maintained between the model's learning state and the threshold as the target domain changes, as shown in eq\ref{5}. Principle \ref{principle3} involves the formulation of class thresholds to modulate the global thresholds, which are then integrated to arrive at the final adaptive thresholds, as shown in eq\ref{6}.

\subsection{Class Prior Alignment}

\begin{wrapfigure}[19]{r}{0.6\textwidth}
\vspace{-25px}
\begin{minipage}{0.55\textwidth}
\begin{algorithm}[H]
    \caption{PLF algorithm at $t$-th iteration.}
    \label{alg}
    \renewcommand{\algorithmicrequire}{\textbf{Input:}}
    \begin{algorithmic}[1]
        \REQUIRE Number of classes $C$, pre-trained model $f_\theta(x)$ ,Unlabeled test data from different domains
        $X^{1:D}$, CAP loss weight $w_c$, EMA decay $\lambda$, and ED factor $\alpha$.
        % \ENSURE A few typical scene data ${\mathcal S_{\rm p}}$  %%output
        \STATE Set the initialize threshold $\tau_0^*(c)$
        (\text{\textcolor{blue}{Eq.}}~\eqref{4})
        
        \FOR{Domain shift $d \in [1,D]$} 
        \STATE Update the global threshold $\tau_t$   (\text{\textcolor{blue}{Eq.}}~~\eqref{5}) 
        
        \STATE Update the class threshold $\tau_t(c)$
        (\text{\textcolor{blue}{Eq.}}~\eqref{6}) 

        \FOR{Class $c \in \mathcal{C}$} 
        
        \STATE $(\tau_t(c),\tau_t) \rightarrow \text{\textcolor{blue}{Eq.}}~\eqref{3} \rightarrow \tau_t^*(c)$
        \ENDFOR
        
        \STATE Compute $\mathcal{L}_u$ on domain unlabeled data 
        (\text{\textcolor{blue}{Eq.}}~\eqref{7})
        
        % \STATE Update histogram $\Tilde{h}$ for $p_t$
        % $(\text{\textcolor{blue}{Eq.}}~\eqref{10})$
        
        \STATE Compute the histogram distribution $U(C)$ and the expectation of probability $\mathbb{E}_B[\cdot]$
        (\text{\textcolor{blue}{Eq.}}~\eqref{10})

        \STATE Compute $\mathcal{L}_c$ on unlabeled data
        (\text{\textcolor{blue}{Eq.}}~\eqref{11})
        
        \RETURN $\mathcal{L} = w_u\mathcal{L}_u + w_c\mathcal{L}_c$ 
        \ENDFOR
    \label{al2}
    \end{algorithmic}
\end{algorithm}
\end{minipage}
\end{wrapfigure}

When conduct Principle \ref{principle3} in the above subsection, we ignore that the diversity of classes, that is, different classes have different domain shift because of different learning difficulties.
% relying only on fine-graining the thresholds is insufficient to address the potentially unbalanced distribution of pseudo-labels generated under domain transfer due to the different learning difficulties of different categories. 
We further propose Class Prior Alignment (CPA), encouraging a more uniform distribution of pseudo-labels across different classes. 
Let the distribution in pseudo-labels be the expectation of model predictions on unlabeled data $X^D$. 
We normalize each prediction $Q$ and $q$ on unlabeled data using the teacher ratio $R_t$ and student ratio $R_s$ between the histogram distribution $U(C)$ and the expectation of probability to calculate the loss weights for each sample to counter the negative effect of imbalance as:

\begin{equation} \label{8}
    R_t = \frac{U(C)_t}{\mathbb{E}_{\max}[Q]} = \frac{{\rm Hist}_B[indi(Q)]}{\mathbb{E}_{\max}[indi(Q)]}.
\end{equation} 
where $indi(\cdot) = \mathbf{1}({{\rm max}(\cdot)}>\tau^*(c)) Q$ represent the probability of filter out pseudo-labels for the teacher model. Similar to $R_t$, we compute $R_s$ as:
\begin{equation}\label{9}
    R_s = \frac{U(C)_s}{\mathbb{E}_{\max}[q]} = \frac{\Tilde{h}}{p},
\end{equation}
where,
\begin{equation} \label{10}
    \Tilde{h} = \lambda \Tilde{h} + (1-\lambda){\rm Hist}_B(q).
\end{equation}
The CPA of loss $L_c$ at the $t$-th iteration is formulated as:
\begin{equation} \label{11}
    \mathcal{L}_{\rm c} = -\mathcal{H}({\rm Normalize}(R_t),{{\rm Normalize}(R_s)}),
\end{equation}
where Normalize = $(\cdot)/\sum\nolimits_{b=1}^B(\cdot)$. The CPA encourages assigning larger weights to predictions with fewer pseudo-labels and smaller weights to predictions with more pseudo-labels, thereby alleviating the imbalance issue.
The overall objective for PLF at the $t$-th iteration is:
\begin{equation} \label{12}
    \mathcal{L} = w_u\mathcal{L}_{\rm u} + w_c\mathcal{L}_{\rm c},
\end{equation}
where $w_{\rm u}$ and $w_{\rm c}$ represent the loss weights of $\mathcal{L}_{\rm u}$ and $\mathcal{L}_{\rm c}$, respectively. Using $\mathcal{L}_{\rm u}$ and $\mathcal{L}_{\rm c}$, PLF balances the information gap of the neural network, thereby reducing classification error.

\begin{table}
\renewcommand{\arraystretch}{1.2}
\centering
\caption{Classification error rate~(\%) on CIFAR10-to-CIFAR10C, CIFAR100-to-CIFAR100C, and ImageNet-to-ImageNet-C.
% The experiments on the highest corruption severity level 5. 
% For CIFAR10C the results are evaluated on WideResNet-28, for CIFAR100C on ResNeXt-29, and for Imagenet-C, ResNet-50 is used. 
We report the performance of our method averaged over 5 runs.
} 
\label{tab:continual-corruptions}
\scalebox{0.8}{
\tabcolsep3pt
\begin{tabular}{l|l|ccccccccccccccc|c}\hline
% & \multicolumn{1}{l|}{Time} & \multicolumn{15}{l|}{$t\xrightarrow{\hspace*{12.2cm}}$}& \\ \hline
% & Method & \rotatebox[origin=c]{70}{Gaussian} & \rotatebox[origin=c]{70}{shot} & \rotatebox[origin=c]{70}{impulse} & \rotatebox[origin=c]{70}{defocus} & \rotatebox[origin=c]{70}{glass} & \rotatebox[origin=c]{70}{motion} & \rotatebox[origin=c]{70}{zoom} & \rotatebox[origin=c]{70}{snow} & \rotatebox[origin=c]{70}{frost} & \rotatebox[origin=c]{70}{fog}  & \rotatebox[origin=c]{70}{brightness} & \rotatebox[origin=c]{70}{contrast} & \rotatebox[origin=c]{70}{elastic} & \rotatebox[origin=c]{70}{pixelate} & \rotatebox[origin=c]{70}{jpeg} & Mean \\
% & Method & Gaussian & shot & impulse & defocus & glass & motion & zoom & snow & frost & fog  & brightness & contrast & elastic & pixelate & jpeg & Mean \\
& Method & Gau. & shot & imp. & def. & glass & mot. & zoom & snow & fro. & fog  & bri. & con. & ela. & pix. & jpeg & Mean \\
\hline
\multirow{8}{*}{\rotatebox[origin=c]{90}{CIFAR10C}}& Source only  & 72.3 & 65.7 & 72.9 & 46.9 & 54.3 & 34.8 & 42.0 & 25.1 & 41.3 & 26.0 & 9.3 & 46.7 & 26.6 & 58.5 & 30.3 & 43.5 \\
& BN      & 28.1 & 26.1 & 36.3 & 12.8 & 35.3 & 14.2 & 12.1 & 17.3 & 17.4 & 15.3 & 8.4 & 12.6 & 23.8 & 19.7 & 27.3 & 20.4 \\
& TENT. & 24.8 & 20.6 & 28.6 &	14.4 & 31.1 & 16.5 & 14.1 & 19.1 & 18.6 & 18.6 & 12.2 & 20.3 & 25.7 & 20.8 & 24.9 & 20.7\\
& Ada.  & 29.1 & 22.5 & 30.0 & 14.0 & 32.7 & 14.1 & 12.0 & 16.6 & 14.9 & 14.4 & 8.1 & 10.0 & 21.9 & 17.7 & 20.0 & 18.5\\
& Cotta      & 24.3 & 21.3 & 26.6 & 11.6 & 27.6 & 12.2 & \textbf{10.3} & 14.8 & 14.1 & 12.4 & \textbf{7.5} & 10.6 & 18.3 & 13.4 & 17.3 & 16.2 \\
 & DSS &24.1&21.3&25.4&11.7&26.9&12.2&10.5&14.5&14.1&12.5&7.8&10.8&18.0&13.1&17.3&
16.0 \\
&PALM&25.9&\text{18.1}&\textbf{22.7}&12.4&25.3&13.2&10.8&13.5&13.2&12.2&8.5&11.9&17.9&12.0&\textbf{15.5}&15.5 \\
& PLF &\textbf{23.5} &18.7 &23.6 &\textbf{10.4}&\textbf{24.4}&\textbf{10.9}&10.6&\textbf{12.7}&\textbf{11.9}&\textbf{10.4}&8.0&\textbf{9.7}&\textbf{16.4}&\textbf{12.0}&16.2&\textbf{14.8} \\
\hline
\multirow{8}{*}{\rotatebox[origin=c]{90}{CIFAR100C}}& Source only  & 73.0&	68.0&	39.4&	29.3&	54.1&	30.8&	28.8&	39.5&	45.8&	50.3&	29.5&	55.1&	37.2	&74.7&	41.2&	46.4\\
& BN  & 42.1 & 40.7 & 42.7 & 27.6 & 41.9 & 29.7 & 27.9 & 34.9 & 35.0 & 41.5 & 26.5 & 30.3 & 35.7 & 32.9 & 41.2 & 35.4 \\
& TENT  & \textbf{37.2} & 35.8 & 41.7 & 37.9 & 51.2 & 48.3 & 48.5 & 58.4 & 63.7 & 71.1 & 70.4 & 82.3 & 88.0 & 88.5 & 90.4 & 60.9 \\
& Ada. & 42.3 & 36.8 & 38.6 & 27.7 & 40.1 & 29.1 & 27.5 & 32.9 & 30.7 & 38.2 & 25.9 & 28.3 & 33.9 & 33.3 & 36.2 & 33.4 \\
& Cotta & 40.1 & 37.7 & 39.7 & 26.9 & 38.0 & 27.9 & 26.4 & 32.8 & 31.8 & 40.3 & 24.7 & 26.9 & 32.5 & 28.3 & 33.5 & 32.5 \\
&DSS &39.7&36.0&37.2&26.3&35.6&27.5&\textbf{25.1}&31.4&30.0&37.8&24.2&26.0&30.0&26.3&31.3&30.9 \\
&PALM&37.3&\textbf{32.5}&\textbf{34.9}&26.2&35.3&27.5&24.6&\textbf{28.9}&29.2&34.1&\textbf{23.5}&27.0&31.1&26.6&34.1&30.2 \\
&PLF &38.2&36.2&37.0&\textbf{25.9}&\textbf{34.7}&\textbf{27.2}&25.6&30.5&\textbf{27.5}&\textbf{32.1}&24.0&\textbf{25.8}&\textbf{27.0}&\textbf{26.0}&\textbf{30.4}&\textbf{29.9} \\
\hline
\multirow{8}{*}{\rotatebox[origin=c]{90}{ImageNet-C}} & Source only  & 97.8 & 97.1 & 98.2 & 81.7 & 89.8 & 85.2 & 78.0 & 83.5 & 77.1 & 75.9 & 41.3 & 94.5 & 82.5 & 79.3 & 68.6 & 82.0 \\
 & BN  & 85.0 & 83.7 & 85.0 & 84.7 & 84.3 & 73.7 & 61.2 & 66.0 & 68.2 & 52.1 & 34.9 & 82.7 & 55.9 & 51.3 & 59.8 & 68.6 \\
 & TENT & 81.6 & 74.6 & 72.7 & 77.6 & 73.8 & 65.5 & \textbf{55.3} & 61.6 & 63.0 & 51.7 & 38.2 & 72.1 & 50.8 & 47.4 & 53.3 & 62.6 \\
 & Ada. & 82.9 & 80.9 & 78.4 & 81.4 & 78.7 & 72.9 & 64.0 & 63.5 & 64.5 & 53.5 & 38.4 & 66.7 & 54.6 & 49.4 & 53.0 & 65.5 \\
 & Cotta  & 84.7 & 82.1 & 80.6 & 81.3 & 79.0 & 68.6 & 57.5 & 60.3 & 60.5 & 48.3 & 36.6 & 66.1 & 47.2 & 41.2 & 46.0 & 62.7 \\
 &DSS &82.3&78.4&76.7&81.9&77.8&66.9&60.9&\textbf{50.8}&60.9&\textbf{47.7}&\textbf{35.4}&69.0&\textbf{47.5}&\textbf{40.9}&46.2&62.2 \\
&PALM&81.1&73.3&70.9&77.0&\textbf{71.9}&62.3&53.9&56.7&60.7&50.4&36.3&65.9&48.1&45.3&48.0&60.1 \\
&PLF &\textbf{78.3}&\textbf{72.4}&\textbf{70.4}&73.8&73.5&\textbf{62.1}&57.4&57.4&\textbf{60.1}&48.9&40.7&\textbf{62.2}&48.5&43.7&\textbf{45.7}&\textbf{59.7}  \\
\hline
\end{tabular}}
  % \vspace{-10px}
\end{table}

\section{Experiments}
\label{Experiments}
\subsection{Set Up}
\textbf{Dataset and Settings.} We conducted extensive experiments to demonstrate the effectiveness of our approach. We evaluate PLF on three benchmark tasks for continual test-time adaptation in image processing: CIFAR-10-C, CIFAR-100-C, and ImageNet-C. These tasks are designed to assess the robustness of machine learning models to corruptions and disturbances in the input data. \textbf{CIFAR10-C} extends the CIFAR-10 dataset, comprising 32 × 32 color images from 10 classes. It includes 15 different corruptions, each at five severity levels, applied to the test images of CIFAR-10 \cite{data}, resulting in a total of 10,000 images. \textbf{CIFAR100-C} extends the CIFAR-100 \cite{data} dataset, containing 32 × 32 color images from 100 classes. It includes 15 different corruptions, each at five severity levels, applied to the test images of CIFAR-100, resulting in 10,000 images in total. \textbf{ImageNet-C} extends the ImageNet \cite{data} dataset, which comprises over 14 million images across more than 20,000 categories. ImageNet-C includes 15 different corruptions, with each corruption having five severity levels. These corruptions are applied to the validation images of ImageNet.
\begin{figure}[t]
  \centering
  \includegraphics[width=\linewidth]{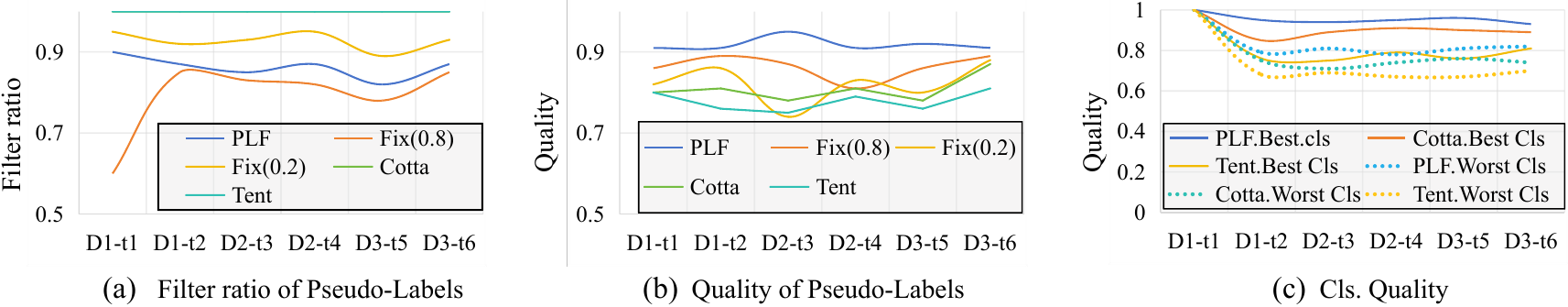}
  \caption{(a): Filter ratio of Pseudo-Label. (b): Quality of Pseudo-Labels. (c): Quality of Pseudo-Labels from the best and worst learned class. Quality is calculated based on the ground truth labels at the underlying level. PLF achieves significantly better performance, see Appendix \ref{Quantity-Quality} for detail.}
  \label{q-q}
\end{figure}

\textbf{Baseline and Implementation details.} We strictly adhere to the CTTA setup, where no source data is accessed. All models, including Cotta, TENT continual, AdaContrast, and DSS, are evaluated online, based on a maximum corruption severity level of five across all datasets. Model predictions are first generated before adapting to the current test stream. Similar to Cotta, we employ standard pre-trained WideResNet \cite{widerestnet}, ResNeXt-29 \cite{restnext-29}, and ResNet-50 \cite{ResNet-50} as the source models for CIFAR10-C, CIFAR100-C, and ImageNet-C. We weigh all loss functions equally using $w_u = w_c = 0.5$.
\subsection{Main Results}

\begin{wraptable}[10]{r}{0.4\textwidth}
\vspace{-15px}
  \centering
  \caption{Analysis of initial threshold}
  % \vspace{-10px}
  \resizebox*{\linewidth}{!}{
\begin{tabular}{l|cccc}
   \toprule
   &Threshold&Gaussian&shot&impulse \\
   \midrule
   \multirow{4}{*}{\rotatebox[origin=c]{90}{low}} 
   &0.10&23.5&18.7&23.6\\   
   &0.15 &23.7 &18.8 &23.7 \\
   &0.20&23.7 &18.9 &23.6 \\
   & 0.25 &23.7&18.9&23.8 \\
\midrule
\multirow{3}{*}{\rotatebox[origin=c]{90}{high}}&0.70&24.1&19.7&24.1\\
&0.80&24.1&19.8&24.2 \\
&0.90&24.2&19.4&24.1\\
\bottomrule
\end{tabular}
}
\label{ini}
\end{wraptable}
\textbf{Results for Continual Test-Time Adaptation.} Table \ref{tab:continual-corruptions} shows the results for each corruption dataset in the continual setting.  Directly testing the source model on target domains in sequence yields high average errors of 43.5\%, 46.4\%, and 77.2\% on CIFAR10-C, CIFAR100-C, and ImageNet-C respectively. Applying BN Stats Adapt \cite{BN1, BN2} to update batch normalization statistics from the current test stream, there is a significant reduction in the average error across all target domains on all datasets. While the TENT-based method aids sequential adaptation to the target domain, it may suffer from severe error accumulation over time. TENT-based method results in a substantially higher error rate of 60.9\% in the long run on CIFAR100-C. Similarly, Conjugate PL performs well in the early adaptation to multiple initial domains but experiences a gradual increase in error rate over time. Conversely, Cotta demonstrates the ability to reduce the average error on most datasets without any signs of error accumulation. However, achieving these results requires heavy test-time augmentation, necessitating up to 32 additional forward passes.

\begin{figure}[t]
  \centering
  \includegraphics[width=\linewidth]{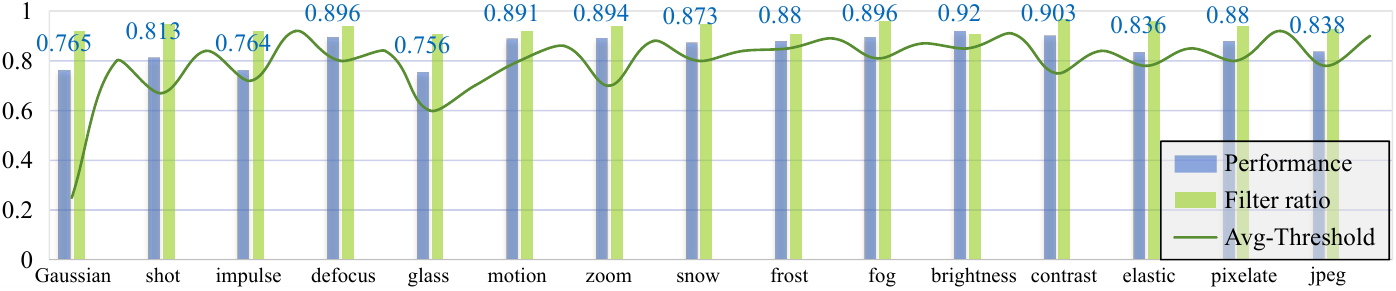}
  \caption{Class threshold and Filter ratio on in CTTA}
  \label{Q-A}
\end{figure}

\begin{wraptable}[9]{r}{0.7\textwidth}
\vspace{-15px}
  \centering
  \caption{Analysis of matching and granularity.}
  % \vspace{10px}
  \resizebox*{\linewidth}{!}{
  \begin{tabular}{lccccc}
   \toprule
    Matching&simple-level&class-level&CIFAR10-C&CIFAR100-C&ImageNet-C\\
   % \cline{4-6}

\midrule
Fixed&\Checkmark&\XSolidBrush&16.8&32.1&63.8\\
DSS&\Checkmark&\XSolidBrush&16.1&31.8&62.2\\
CRM+ED&\Checkmark&\XSolidBrush&16.2&31.7&62.8\\
EMA+ED&\Checkmark&\XSolidBrush&15.9&31.4&61.9 \\
EMA+ED&\XSolidBrush&\Checkmark&15.4&29.9&59.7 \\
   \bottomrule
\end{tabular}
}
\label{Matching and Granularity}
% \vspace{-15px}
\end{wraptable}
By utilizing adaptive thresholding for pseudo-label filtering, our proposed PLF consistently outperforms Cotta across all datasets. This indicates that PLF contributes to better adaptation of the model to 
continuous target domains, reducing the impact of error accumulation caused by noisy pseudo-labels. We observe that this  phenomenon becomes more pronounced in more challenging datasets, where the model exhibits lower certainty in continuing to the target domain.Compared to Cotta, PLF successfully reduces the average errors on CIFAR100-C and ImageNet-C from 32.5\% to 30.1\% and from 66.8\% to 59.4\%. To further investigate the effectiveness of PLF over the baseline, we also evaluate its adaptation performance over ten different sequences on ImageNet-C. There is a 3\% decrease on average over 10 diverse sequences in comparison with Cotta, indicating that our method is more robust to the order of the target domain sequence. 

\textbf{For Testing On Principles.}
Take CIFAR10-C for example. Table.\ref{ini} shows the advantage of low initial thresholds in the early target domain. Second, we test for different threshold and model confidence matching methods, including Fixed, DSS, Confidence Ratio Matching (CRM) \cite{CRM}+ED, and EMA+ED. Table.\ref{Matching and Granularity} shows that positive correlation matching methods generally perform better. Finally, we compared the category granularity, as shown in Table.\ref{Matching and Granularity}, fine-grained thresholding can distinguish similar pseudo-labels more effectively.

\begin{table}[h]
  % \vspace{-10px}
  \centering
  \caption{Analysis of Pseudo-Label Filtering.}
  % \vspace{-10px}
  \resizebox*{\linewidth}{!}{
\begin{tabular}{lccc|ccc|ccc}
   \toprule
   Avg. Error (\%)&CIFAR10-C&Filter ratio&Quality&CIFAR100-C&Filter ratio&Quality&ImageNet-C&Filter ratio&Quality \\
   \midrule
   Cotta &16.2 &-&0.87&32.5 &-&0.81&62.7 &-&0.71\\
   Cotta(w/ Filtering) &15.6 &0.90 &0.91 &31.2 &0.87&0.86&61.9&0.84&0.81 \\
   \midrule
   TENT &20.7&-&0.84 &60.9&-&0.69 &62.6&-&0.61 \\
   TENT(w/ Filtering) &19.4&0.88&0.87 &57.1 &0.85 &0.78 &61.4&0.81 &0.68\\
   \midrule
   PLF(w/ fixed $\tau@0.8$ ) &15.9&0.89&\textbf{0.94} &32.1&0.82&0.89 &62.1&0.78&0.82 \\
   PLF(w/ Filtering $\tau$) &\textbf{14.7}&\textbf{0.91}&0.92&\textbf{30.1}&\textbf{0.89}&\textbf{0.91}&\textbf{59.7}&\textbf{0.86}&\textbf{0.86} \\
   % \cline{4-6}
   \bottomrule
\end{tabular}
}
\vspace{-10px}
\label{Threshold}
\end{table}

\textbf{PLF for Continual Test-Time Adaptation.}
As shown in Table \ref{Threshold}, to observe the advantages of the PLF method in pseudo-label filtering, we conducted experiments to evaluate the filter ratio and quality of pseudo-labels for methods Cotta(w/filtering), TENT(w/filtering), and PLF. The results indicate that not filtering pseudo-labels (Cotta, TENT) leads to a significant accumulation of low-quality pseudo-labels, resulting in error accumulation. While using a fixed threshold method for pseudo-label filtering can maintain quality to some extent, it may incorrectly filter out correct pseudo-labels, leading to incomplete learning of domain knowledge by the model. In contrast, PLF, employing adaptive thresholds, can maintain relatively high quality on a high filter ratio of pseudo-labels. Finally, Fig.\ref{q-q} shows the trend of pseudo-labels' filter ratio and quality for the pseudo-labeling method, and Fig.\ref {Q-A} illustrates the changing trends of average class thresholds and  Filter ratio in CIFAR10-C. It can be observed that the adaptive threshold varies with the transformation of the domain, which ensures the stability of the filtering rate so that the model can converge with high quality.

\begin{wraptable}[12]{r}{0.6\textwidth}
\vspace{-15px}
  \centering
  \caption{Ablation studies of intra-scene.}
  % \vspace{10px}
  \resizebox*{\linewidth}{!}{
  \begin{tabular}{ccccccc}
   \toprule
    Init&Fixed&SAT&CPA&CIFAR10-C&CIFAR100-C&ImageNet-C\\
   % \cline{4-6}
   \midrule
   0.8&\Checkmark&\XSolidBrush&\XSolidBrush  &15.9&32.1&62.1\\
    0.8&\Checkmark&\XSolidBrush&\Checkmark  &15.5&31.7&61.8\\
   0.8&\XSolidBrush &\Checkmark &\XSolidBrush & 15.1&30.9&60.7 \\
   0.8&\XSolidBrush &\Checkmark &\Checkmark &15.1&30.4&60.1 \\
\midrule
  $\frac{1}{C}$&\Checkmark&\XSolidBrush&\XSolidBrush &16.2&32.1&62.4\\
$\frac{1}{C}$&\Checkmark&\XSolidBrush&\Checkmark  &15.8&31.4&61.9\\
 $\frac{1}{C}$&\XSolidBrush &\Checkmark &\XSolidBrush &15.2&30.8&60.3 \\
$\frac{1}{C}$&\XSolidBrush &\Checkmark &\Checkmark &14.8&29.9&59.7 \\
   \bottomrule
\end{tabular}
}
\label{Ablation}
% \vspace{-15px}
\end{wraptable}
\textbf{Ablations.} In the ablation experiments, as shown in Table\ref{Ablation}. Firstly, we analyze different initial thresholds. We found that under fixed threshold conditions, higher thresholds bring significant benefits, as fixed lower thresholds may lead to excessive learning of low-quality samples. However, under dynamic threshold conditions, initial lower thresholds can have certain advantages.
Next, we compared the two states of whether to use adaptive thresholds. We found that the dynamic growth of thresholds can filter out low-quality pseudo-labels, thereby optimizing the model. Therefore, we believe that the adaptive threshold method achieves a better balance between maximizing target adaptability and minimizing error accumulation.
Finally, we demonstrated the effectiveness of the CPA method. By adjusting the class distribution to ensure fairness in classification, it helps the model learn and diversify better. This method effectively addresses the impact of adversarial imbalanced base distributions and accurately reflects changes in sample distributions brought about by domain shifts or underlying batch data differences.

\section{Conclusion and Limitations}
\label{Conclusion}
In this paper, we introduce a novel approach, referred to as PLF, for the CTTA task, based on adaptive thresholds and the CPA method. To address the challenge of error accumulation in traditional CTTA methods, we continuously monitor the confidence of predictions online, thereby setting adaptive thresholds to select high-quality samples for different testing strategies. Additionally, we align class distributions to ensure fairness in classification. We believe that confidence thresholding has greater potential in CTTA, but there are potential \textbf{limitations}: First, the role of an initially low threshold acts only with forward continuous domains, and as the CTTA process lengthens, the more limited that role becomes. Second, the adaptive thresholds we designed are somewhat weak for more complex datasets, and we conjecture that under unsupervised conditions, the more categories there are, the less the PLF will learn about category confidence, leading to a reduced role of pseudo-label filtering. Finally, the model's fitness still relies on the confidence predictions of the classifiers, especially in the presence of destructive domain changes. We hope that the effective performance of the PLF will inspire further research on optimal thresholding.

% \section*{References}

% References follow the acknowledgments in the camera-ready paper. Use unnumbered first-level heading for
% the references. Any choice of citation style is acceptable as long as you are \cite{53}
% consistent. It is permissible to reduce the font size to \verb+small+ (9 point)
% when listing the references.
% Note that the Reference section does not count towards the page limit.
% \medskip
{
\small

\bibliographystyle{plainnat}
% \bibliography{ref}
}

\clearpage
\appendix

\section{Appendix / supplemental material}
\subsection{Proof of Lemma 3.1}
\label{proof}
\textbf{Lemma 3.1 (Class Probability Distribution (CPD))}:
For a multi-class classification problem as mentioned above, the pseudo label $Y_p$ has the following probability distribution:
\begin{equation}
\begin{aligned}
& P(X|Y_p = c) = \Phi\left(\frac{1}{\sigma_c}\left(\mu_c-\frac{1}{\beta}\log(\frac{\tau}{1-\tau})-\frac{1}{\beta}\log\sum\nolimits_{i \neq c} e^{\beta l_i}\right)\right), \\
& P(X|Y_p = 0) = 1-\sum\nolimits_{i}P({X|Y_p} = i),
\end{aligned}
\nonumber
\end{equation}
where $\Phi$ is the cumulative distribution function of a standard normal distribution.

Since multi-classification calculations are too complex, Lemma 3.1 Let's take a binary classification problem as an example. Pseudo-label $Y_p$ has the following probability distribution:
\begin{equation}
X | Y = -1 \sim N(\mu_1,\sigma_1^2) , X | Y = +1 \sim N(\mu_2,\sigma_2^2).
\nonumber
\end{equation}

Assuming without loss of generality that $\mu_1 > \mu_2$, the confidence score s(x) output by the classifier can be expressed as $s(x) = 1/[1 + exp(- \beta (x-(\mu_1 + \mu_2 )/2))]$, where $\beta$ is a positive parameter reflecting the model's learning state. The model's confidence is influenced by continuous domain changes, exhibiting a stepped growth rather than a smooth one. $(\mu_1 + \mu_2)/2$ serves as the optimal linear decision boundary in Bayesian terms. We consider a scenario where a fixed threshold $\tau \in (1/2,1)$ is used to generate pseudo-labels. If $s(x) > \tau$, the sample x is assigned the pseudo-label +1; if $s(x) < 1 - \tau$, the pseudo-label is -1. If $1 - \tau \leq s(x) \leq  \tau$, the pseudo-label is 0 (masked).

\begin{equation}
\begin{aligned}
& P({X|Y_p} = 1) = \frac{1}{2}\Phi (\frac{{\frac{{{\mu _2} - {\mu _1}}}{2} - \frac{1}{\beta}\log (\frac{{{\tau}}}{{1 - {\tau}}})}}{{{\sigma _2}}}) + \frac{1}{2}\Phi (\frac{{\frac{{{\mu _1} - {\mu _2}}}{2} - \frac{1}{\beta}\log (\frac{{{\tau}}}{{1 - {\tau}}})}}{{{\sigma _1}}}),\\
& P(X|{Y_p} = -1) = \frac{1}{2}\Phi (\frac{{\frac{{{\mu _2} - {\mu _1}}}{2} - \frac{1}{\beta}\log (\frac{{{\tau}}}{{1 - {\tau}}})}}{{{\sigma _1}}}) + \frac{1}{2}\Phi (\frac{{\frac{{{\mu _1} - {\mu _2}}}{2} - \frac{1}{\beta}\log (\frac{{{\tau}}}{{1 - {\tau}}})}}{{{\sigma _2}}}), \\
& P(X|{Y_p} = 0) = 1-P({Y} = 1) -P({Y} = -1) ,
\end{aligned}
\nonumber
\end{equation}

where $\Phi$ is the cumulative distribution function of a standard normal distribution. Moreover, $P({Y} = 0)$ increases as ${\mu _2} - {\mu _1}$ gets smaller.

\begin{proof}
\label{proof-1}

sample x will be assigned pseudo label 1 if
\begin{equation}
    \frac{1}{1 + exp(-(x-(\frac{\mu_1 + \mu_2}{2}))} > \tau,
\nonumber
\end{equation}
which is equivalent to
\begin{equation}
    x > \frac{\mu_1 + \mu_2}{2} + \frac{1}{\beta}\log(\frac{\tau}{1-\tau}),
    \nonumber
\end{equation}
Likewise, x will be assigned pseudo label -1 if
\begin{equation}
    1/[1 + exp(-(x-(\mu_1 + \mu_2 )/2))] < 1-\tau,
    \nonumber
\end{equation}
which is equivalent to
\begin{equation}
    x < \frac{\mu_1 + \mu_2}{2} - \frac{1}{\beta}\log(\frac{\tau}{1-\tau}),
    \nonumber
\end{equation}

Remembering that the probabilities of $P(Y = 1) = P(Y = -1) = 0.5$, when we perform integration over x, we obtain the following conditional probabilities:
% \begin{equation}
% \begin{aligned}
%     P(Y = 1|Y_t=1) = \Phi(\frac{\frac{\mu_2 - \mu_1}{2}-\log(\frac{\tau}{1-\tau})}{\sigma _2}) \\
%     P(Y = 1|Y_t=-1) = \Phi(\frac{\frac{\mu_1 - \mu_2}{2}-\log(\frac{\tau}{1-\tau})}{\sigma _1}) \\
%     P(Y = -1|Y_t=-1) = \Phi(\frac{\frac{\mu_2 - \mu_1}{2}-\log(\frac{\tau}{1-\tau})}{\sigma _1}) \\
%      P(Y = -1|Y_t=1) = \Phi(\frac{\frac{\mu_1 - \mu_2}{2}-\log(\frac{\tau}{1-\tau})}{\sigma _2})
% \end{aligned}
%     \nonumber
% \end{equation}

\begin{equation}
\begin{aligned}
& P({Y_p} = 1) = \frac{1}{2}\Phi (\frac{{\frac{{{\mu _2} - {\mu _1}}}{2} - \frac{1}{\beta}\log (\frac{{{\tau}}}{{1 - {\tau}}})}}{{{\sigma _2}}}) + \frac{1}{2}\Phi (\frac{{\frac{{{\mu _1} - {\mu _2}}}{2} - \frac{1}{\beta}\log (\frac{{{\tau}}}{{1 - {\tau}}})}}{{{\sigma _1}}}),\\
& P({Y_p} = -1) = \frac{1}{2}\Phi (\frac{{\frac{{{\mu _2} - {\mu _1}}}{2} - \frac{1}{\beta}\log (\frac{{{\tau}}}{{1 - {\tau}}})}}{{{\sigma _1}}}) + \frac{1}{2}\Phi (\frac{{\frac{{{\mu _1} - {\mu _2}}}{2} - \frac{1}{\beta}\log (\frac{{{\tau}}}{{1 - {\tau}}})}}{{{\sigma _2}}}), \\
\end{aligned}
\nonumber
\end{equation}

\end{proof} 
\subsection{Derivation of Principle-2}
\label{Derivation2}
To show that $P(Y_p = 0)$ increases as $\tau$ gets bigger or $\beta$ gets smaller , we only need to show $P(Y_p = 1 || -1)$ gets smaller. First, we represent the irrelevant variables as constants, namely $a = \frac{\mu_2-\mu_1}{2}, b_1 = \frac{1}{2\sigma_1},b_2 = \frac{1}{2\sigma_2},z =\frac{1}{\beta} $
Then, we denote $\log(\frac{\tau}{1-\tau})$ as $c$. Finally, due to the symmetry of $P(Y = 1)$ and $P(Y_p = -1)$\{$b_1 = b_2$\} we obtain:
\begin{equation}
    P(Y_p =1) = P(Y_p = -1) = \frac{1}{2}\phi(ab - bcz) + \frac{1}{2}\Phi(-ab-bcz),
    \nonumber
\end{equation}
Derivatives for $c$ and $z$ are taken respectively, we have:
\begin{equation}
\begin{aligned}
    \\ f'(c) = -\frac{1}{2}b(\phi(ab-bc)-\phi(-ab-bc)),
    \\ h'(z) = -\frac{1}{2}bc(\Phi(ab-bcz)-\phi(-ab-bcz),
\end{aligned}
\nonumber
\end{equation}
where $\phi$ is the probability density function of a standard normal distribution. Since$|ab-bcz|<|-ab-bcz| , |ab-bcz|<|-ab-bcz|$, we have $f'(c) < 0 , h'(z) < 0$, and the proof-1 is complete.

\subsection{Derivation of Principle-3}
\label{Derivation3}
Similarly, to show that $P(Y_p = 0)$ increases as $z = \mu_2 - \mu_1$ gets smaller, we only need to show $P(Y_p = 1 || -1)$ gets bigger. First, we represent the irrelevant variables as constants, namely $n_1 = \frac{1}{2\sigma_1}, n_2 = \frac{1}{2\sigma_2}, m_2 = \frac{\log(\frac{\tau}{1-\tau})}{\sigma_1} $
Then, we denote $\mu_2 - \mu_1$ as $z$. Finally, due to the symmetry of $P(Y_p = 1)$ and $P(Y_p = -1)$
\{($n = n_1 = n_2,m = m_1 = m_2$)\} we obtain:
\begin{equation}
    P(Y_p =1) = P(Y_p = -1) = \frac{1}{2}\Phi(nz-m) +  \frac{1}{2}\Phi(-nz-m),
    \nonumber
\end{equation}
take the derivative of $z$, we have:
\begin{equation}
    g'(z) = \frac{1}{2}a(\phi(nz-m)) - \phi(-nz-m)),
    \nonumber
\end{equation}
where $\phi$is the probability density function of a standard normal distribution. Since$|nz-m|<|-nz-m|$,we have $g'(z) > 0$, and the proof-2 is complete.

Based on the above proof, two conclusions can be drawn. Firstly, during domain adaptation, there exists an inverse relationship between the threshold and the model confidence. As the model confidence abruptly decreases during domain adaptation, resulting in a reduction in the sampling rate, it is necessary to lower the threshold to ensure a stable sampling rate. Additionally, expanding the inter-class distribution can also increase and stabilize the sampling rate.

\section{Filter ratio and Quality Trade-off}
\label{Quantity-Quality}
In this section, we will elaborate on the definition of filter ratio and quality formulas and their derivation process. Given that the threshold filtering method will have an impact on the filter ratio and quality of pseudo-labeling, we will comprehensively consider these two factors and pseudo-labeling technology to design the corresponding evaluation index.

\subsection{Filter ratio}
The definition and derivation of filter ratio $f(p)$ of pseudo-labels is rather straightforward. We define the filter ratio as the percentage/ratio of unlabeled data enrolled in the weighted unsupervised loss. In other words, the filter ratio is the average sample weights on unlabeled data:
\begin{equation}
    f(p) = \frac{\sum\nolimits_{n=1}^{N} {\mathbf{1}({\rm max}(q)>\tau)}}{N}.
\label{Quantity}
\nonumber
\end{equation}

\subsection{Quality}
We define the quality $g(p)$ of pseudo-labels as the percentage/ratio of correct pseudo-labels enrolled in the weighted unsupervised loss, assuming the ground truth label $Y$ of unlabeled data is known. With the $0/1$ correct indicator function $\gamma(p)$ being defined as:
\begin{equation}
    \gamma(p) = \mathbf{1}(({\rm max}(q)>\tau) || \hat{p} = Y) \in \{0,1\},
    \nonumber
\end{equation}
where $\hat{p}$ is the one-hot vector of pseudo-label argmax(p). We can formulate quality as:
\begin{equation}
    g(p) = \frac{\mathbf{1}(({\rm max}(q)>\tau) || \hat{p} = Y)}{\mathbf{1}(({\rm max}(q)>\tau)}.
\label{Quality}
\nonumber
\end{equation}

% \begin{table}[h]
% \centering
% \caption{Semantic segmentation results (mIoU in \%) on the Cityscapes-to-ACDC online continual test-time adaptation task. }
% \label{tab:acdc}
% \scalebox{0.75}{
% \tabcolsep3pt
% \begin{tabular}{l|cccc|cccc|cccc|cccc|c}\hline
% % Time &  \multicolumn{16}{l|}{$t\xrightarrow{\hspace*{14cm}}$}\\ \hline
% Round      & \multicolumn{1}{l}{1} & \multicolumn{1}{l}{} & \multicolumn{1}{l}{} & \multicolumn{1}{l|}{} & \multicolumn{1}{l}{4} & \multicolumn{1}{l}{} & \multicolumn{1}{l}{} & \multicolumn{1}{l|}{} & \multicolumn{1}{l}{7} & \multicolumn{1}{l}{} & \multicolumn{1}{l}{} & \multicolumn{1}{l|}{} & \multicolumn{1}{l}{10} & \multicolumn{1}{l}{} & \multicolumn{1}{l}{} & \multicolumn{1}{l|}{} & \multicolumn{1}{l}{All} \\ \hline
% Condition& Fog  & Night& Rain  & Snow& Fog  & Night & Rain& Snow & Fog& Night & Rain & Snow & Fog & Night & Rain& Snow  & Mean  \\ 
% \hline
% Source & 69.1  & 40.3   & 59.7   & 57.8  & 69.1    & 40.3      & 59.7  & 57.8   & 69.1   & 40.3  & 59.7   & 57.8        & 69.1   & 40.3   & 59.7   & 57.8   & 56.7   \\
% BN Stats Adapt & 62.3 & 38.0 & 54.6 & 53.0& 62.3 & 38.0 & 54.6 & 53.0& 62.3 & 38.0 & 54.6 & 53.0& 62.3 & 38.0 & 54.6 & 53.0 & 52.0\\
% TENT-continual~\cite{}  & 69.0                  & 40.2   & 60.1   & 57.3   & 66.5  & 36.3  & 58.7           & 54.0  & 64.2  & 32.8 & 55.3  & 50.9 & 61.8& 29.8& 51.9& 47.8 & 52.3 \\ 
% Cotta & 70.9& 41.2& 62.4&59.7& 70.9&41.0& 62.7&59.7& 70.9& 	41.0& 62.8& 59.7& 70.8& 41.0& 62.8&59.7& 58.6 \\ \hline
   
% \end{tabular}}
% \end{table}
\section{Results for Gradual Test-Time Adaptation.}
As shown in Table \ref{tab:gradual-corruptions}, our results in the CIFAR10-CIFAR10C progressive task exhibit a 2.4\% improvement in error compared to the state-of-the-art (SOTA) Cotta method. CIFAR100-to-CIFAR100C poses a more challenging task due to its larger number of categories compared to CIFAR10C. Surprisingly, our method still outperforms the Cotta method by 2\% on CIFAR100C. It maintains strong performance across various blur types, indicating the continued advantage of using adaptive thresholding through inter-class relationship modulation and balanced prediction optimization across classes. To further demonstrate the effectiveness of our proposed method on a broader range of datasets, we conducted experiments on ImageNet to ImageNet-C. Our method achieves the best performance, surpassing the Cotta method by 2\%.
\begin{table}[h]
\renewcommand{\arraystretch}{1.1}
\centering
\caption{Classification error rate~(\%) for the gradual CIFAR10-to-CIFAR10C, CIFAR100-to-CIFAR100C, and ImageNet-to-ImageNet-C benchmark averaged over all 15 corruptions. We separately report the performance averaged over all severity levels (@ level 1--5) and averaged only over the highest severity level 5 (@ level 5). The number in brackets denotes the difference to the continual benchmark.}

\label{tab:gradual-corruptions}
\scalebox{0.93}{
\tabcolsep3pt
\begin{tabular}{l|l|ccccccc}\hline
& & Source & BN & TENT & AdaCont. & Cotta &DSS &PLF  \\ \hline
\multirow{2}{*}{CIFAR10C} & @level 1--5  & 24.7 & 13.7 & 20.4 & 12.1 & 10.9 &9.8 &\textbf{8.5} \\
& @level 5 & 43.5 & 20.4 & 25.1 \small{(+4.4)} & 15.8 \small{(-2.7)} & 14.2 \small{(-2.0)} & 12.9 \small{(-3.1)} & \textbf{10.1} \small{(-3.8)} \\ 
\hline
\multirow{2}{*}{CIFAR100C} & @level 1--5 & 33.6 & 29.9 & 74.8 & 33.0 & 26.3 &26.6 &\textbf{25.8}  \\
& @level 5 & 46.4 & 35.4 & 75.9 \small{(+15.0)} & 35.9 \small{(+2.5)} & 28.3 \small{(-4.2)} & 28.5 \small{(-3.8)} & \textbf{26.7} \small{(-2.3)} \\ \hline
\multirow{2}{*}{ImageNet-C} & @level 1--5  & 58.4 & 48.3 & 46.4 & 66.3&\textbf{38.8} &39.5 &39.1 \\
& @level 5 & 82.0 & 68.6 & 58.9 \small{(-3.7)} & 72.6 \small{(+7.1)} & 43.1 \small{(-19.6)} & 44.9 \small{(+7.5)} & \textbf{41.1} \small{(-17.6)}\\ \hline
\end{tabular}}
\end{table}

\section{Analysis of Initial Threshold}
\label{Initial}
% \subsection{Low
% and High thresholds on CTTA experiments}
% Since the initial threshold is more useful for the starting target domain, the reason is that high thresholds do not affect the error accumulation problem too much, although they incorrectly filter some high-quality pseudo-labels. We therefore compared the effects of multiple low thresholds as well as high thresholds on CTTA experiments in the first three domains on the CIFAIR-10C dataset. 

% \begin{table}[h]
%   \centering
%   \caption{Analysis of initial threshold}
%   % \vspace{-10px}
%   \resizebox*{0.5\linewidth}{!}{
% \begin{tabular}{l|cccc}
%    \toprule
%    &Threshold&Gaussian&shot&impulse \\
%    \midrule
%    \multirow{4}{*}{\rotatebox[origin=c]{90}{low}} 
%    &0.10&23.5&18.7&23.6\\   
%    &0.15 &23.7 &18.8 &23.7 \\
%    &0.20&23.7 &18.9 &23.6 \\
%    & 0.25 &23.7&18.9&23.8 \\
% \midrule
% \multirow{3}{*}{\rotatebox[origin=c]{90}{high}}&0.70&24.1&19.7&24.1\\
% &0.80&24.1&19.8&24.2 \\
% &0.90&24.2&19.4&24.1\\
% \bottomrule
% \end{tabular}
% }
% \vspace{-10px}
% \label{ini}
% \end{table}

\subsection{Initialization Threshold}
\label{initialization}
There should be an optimal initial low threshold for different datasets, based on Principle \ref{principle3}, the first thing that comes to our mind is the number of categories in the dataset, therefore, we derive the threshold by the ideal sampling rate, and the results show that it is indeed related to the number of categories. For a multi-class classification problem as mentioned above, the pseudo label $Y_p$ has the following probability distribution:
\begin{equation}
P(X|Y_p = c) = \Phi\left(\frac{1}{\sigma_c}\left(\mu_c-\frac{1}{\beta}\log(\frac{\tau}{1-\tau})-\frac{1}{\beta}\log\sum\nolimits_{i \neq c} e^{\beta l_i}\right)\right), 
\nonumber
\end{equation}
The theoretical basis for the initialization threshold $\tau_0(c)$ of $1/C$ is derived in detail in this chapter. Due to the poor state of the initial model, it is necessary to test it using a large number of samples to converge the model quickly. Therefore, we assume a sampling rate of $1$. The following is the proof procedure:
\begin{equation}
1 = \Phi\left(\frac{1}{\sigma_c}\left(\mu_c-\frac{1}{\beta}\log(\frac{\tau}{1-\tau})-\frac{1}{\beta}\log\sum\nolimits_{i \neq c} e^{\beta l_i}\right)\right),
\nonumber
\end{equation}
which is equivalent to
\begin{equation}
    \frac{1}{\sigma_c}\left(\mu_c-\frac{1}{\beta}\log(\frac{\tau}{1-\tau})-\frac{1}{\beta}\log\sum\nolimits_{i \neq c} e^{\beta l_i}\right) = 0.
\nonumber
\end{equation}
Calculation Threshold to
\begin{equation}
\tau = \frac{1}{\sum\nolimits_{i \neq c} e^{\beta l_i}/e^{\beta\mu_c}+1},
\nonumber
\end{equation}
where, Simply observe whether the order of magnitude of $\sum\nolimits_{i \neq c} e^{\beta l_i}/e^{\beta\mu_c}+1$ is related to the number of categories:
\begin{equation}
    \sum\nolimits_{i \neq c} e^{\beta l_i}/e^{\beta\mu_c}+1 \approx \frac{k}{C} +1,
\nonumber
\end{equation}
where $k$ represents the scaling factor between categories.

\section{Comparison Experiment of Positive Correlation}
\label{pos}
\begin{figure}[h]
  \centering
  \includegraphics[width=\linewidth]{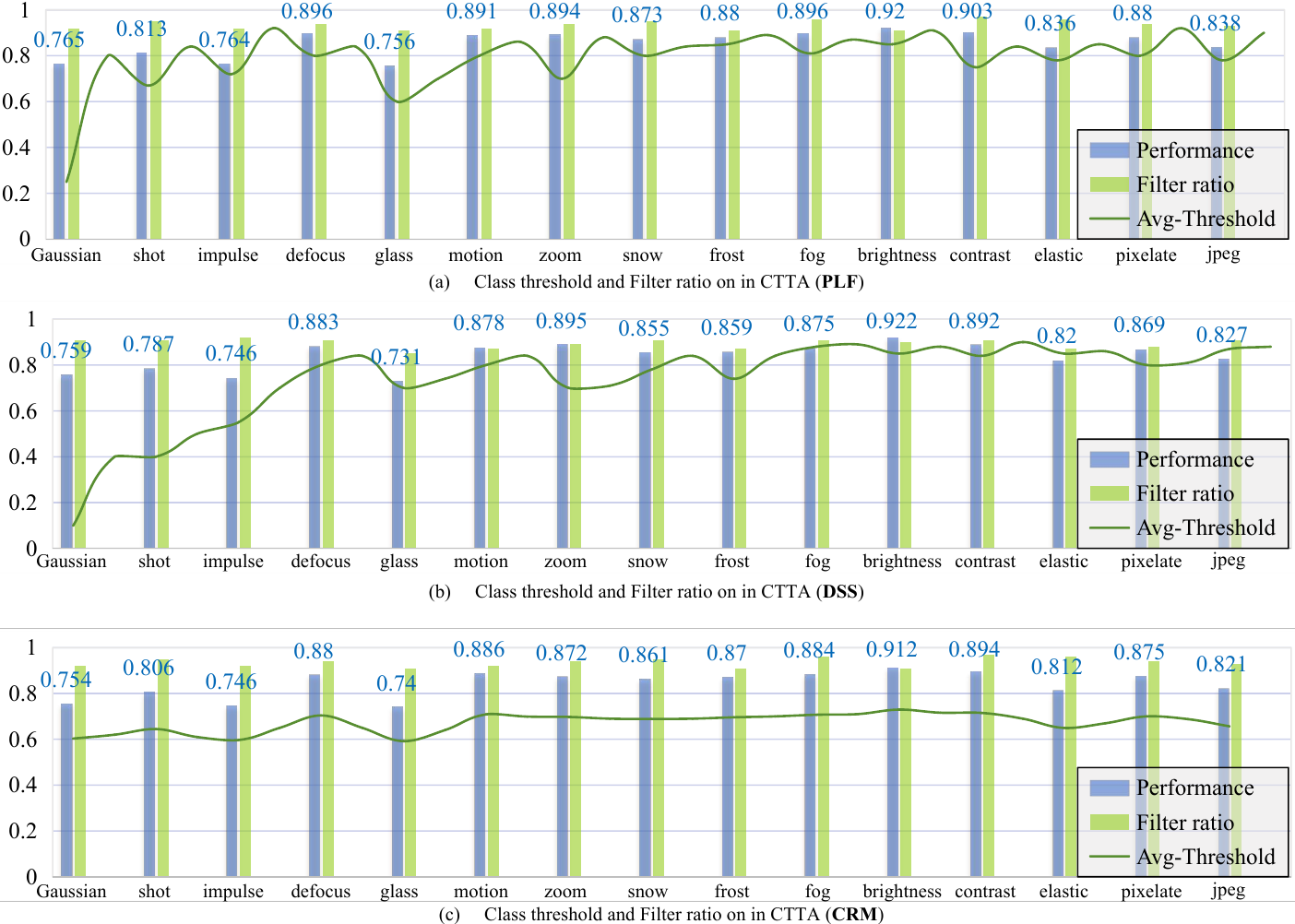}
  \caption{Trends in Avg-Threshold and filter ratios in the CIFAR10-C dataset under the three maintenance of positive correlation algorithms.}
  \label{positive}
  % \vspace{-15px}
\end{figure}

The purpose of this section is to illustrate, through a comparative experimental analysis, as shown in Fig. \ref{positive}, the reasons for the EMA and ED algorithms used in this paper compared to other algorithms that can maintain a positive correlation between thresholds and model confidence.
By comparing with the DSS method and the Confidence Ratio Matching (CRM) method, we can find that using our proposed adaptive thresholding method, the pseudo-labels can be filtered better, which improves the classification performance of the model in the target domain. We further analyze the algorithms and find that the use of the EMA and ED algorithms allows the thresholds to not only be positively correlated with the model confidence but also to combine the model confidence of the past time steps for the purpose of initially predicting the thresholds for the next time step.

\section{Extend analysis on Class Prior Alignment}
\begin{figure}[h]
  \centering
  \includegraphics[width=\linewidth]{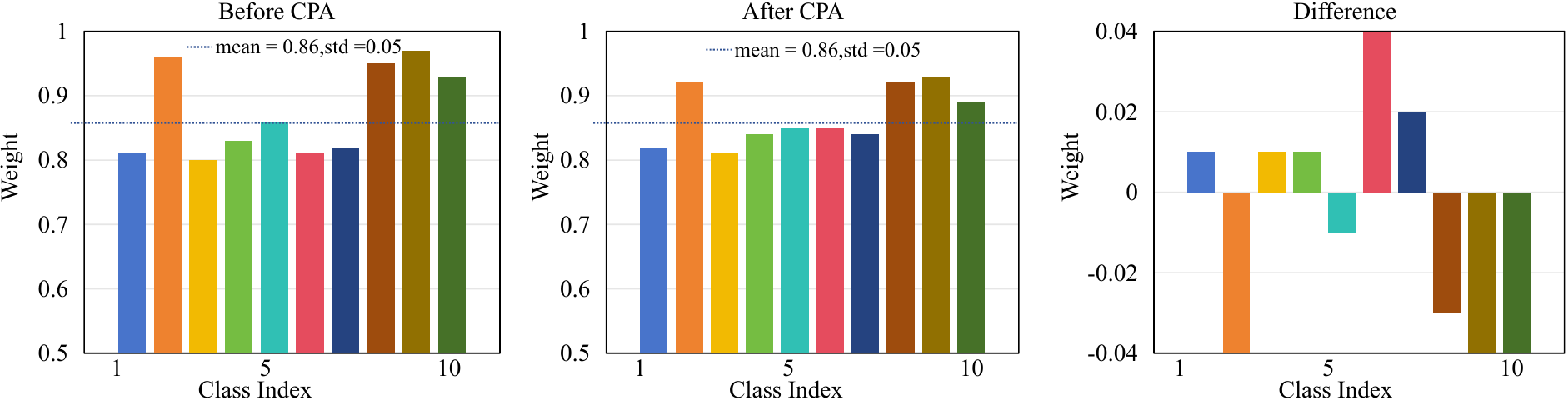}
  \caption{Average weight for each class according to pseudo-label, for (a) before CPA; and (b) after CPA. We also include the difference between them in (c).}
  \label{CPA}
  % \vspace{-15px}
\end{figure}

In this section, we provide more explanation regarding the mechanism of Class Prior Alignment (CPA). CPA is proposed to make the model learn more equally in each class to reduce the pseudo-label imbalance. To visualize this, we plot the average class weight according to pseudo-labels of PLF before CPA and after CPA at the beginning of testing, as shown in Fig. \ref{CPA} facilitates a more balanced class-wise sample weight, which would help the model learn more equally on each class.

\section{Experience Detail}
\label{Experience Detail}
The hyperparameters for image classification evaluation are shown in Table \ref{Hyper-parameters}. We use the Adam optimizer instead. For a more similar comparison with SOTA, WideResNet is for CIFAR-10C, ResNeXt-29 is for CIFAR-100C, and ResNet-50 is for ImageNetC. Use NVIDIA V100 to test image classification.
\begin{table}[h]
  \centering
  \caption{Hyper-parameters of long-tailed image classification tasks.}
  % \vspace{-10px}
  \label{Hyper-parameters}
  \resizebox*{0.9\linewidth}{!}{
  \begin{tabular}{c|ccc}
   \toprule
 Dataset &CIFAR-10C &CIFAR-100C &ImageNetC \\
 \midrule
 Model &WideResNet &ResNeXt-29 &ResNet-50 \\
  \midrule
 Batch size &200 &200 &200 \\
  \midrule
 Learning Rate &\multicolumn{3}{c}{0.01} \\
 Optimizer &\multicolumn{3}{c}{Adam} \\
 Model EMA Momentum &\multicolumn{3}{c}{0.9} \\
 Weak Augmentation &\multicolumn{3}{c}{Student prediction} \\
 Strong Augmentation Augmentation &\multicolumn{3}{c}{Teacher prediction (RandAugment)} \\
 Exponential decay factor &\multicolumn{3}{c}{0.4}\\
   \bottomrule
   \end{tabular}
}
\vspace{15px}
\end{table}

\end{document}